\pdfoutput=1

\documentclass[11pt]{article}

\usepackage[preprint]{latex/acl}

\usepackage{times}
\usepackage{latexsym}

\usepackage[T1]{fontenc}

\usepackage[utf8]{inputenc}

\usepackage{microtype}

\usepackage{inconsolata}

\usepackage{graphicx}

\usepackage{latex/ai-usage-card} 
\usepackage{latex/annotation}
\usepackage{amsmath}
\usepackage{stfloats}
\usepackage{multirow}
\usepackage{makecell}
\usepackage{arydshln}
\usepackage{booktabs}
\usepackage{amssymb}
\usepackage{enumitem}
\usepackage{cleveref}
\usepackage{tcolorbox}
\newcounter{mysub}

\crefalias{mysub}{subsection}
\usepackage{needspace}

\title{Multi-Agent Reasoning Improves Compute Efficiency:\\ Pareto-Optimal Test-Time Scaling}

\author{
 \textbf{Florian Valentin Wunderlich,}
 \textbf{Lars Benedikt Kaesberg,}
 \textbf{Jan Philip Wahle,}
\\
\textbf{Terry Ruas,}
 \textbf{Bela Gipp}
\\
\\
 University of Göttingen, Germany
\\
 \small{
   \textbf{Correspondence:} \href{mailto:florian.wunderlich@uni-goettingen.de}{florian.wunderlich@uni-goettingen.de}
 }
}

\begin{document}
\maketitle
\AddAnnotationRef{}

\begin{abstract}
Advances in inference methods have enabled language models to improve their predictions without additional training. 
These methods often prioritize raw performance over cost-effective compute usage. 
However, computational efficiency is key for real-world applications with resource constraints. 
We provide a systematic analysis of the inference scaling strategies \textit{self-consistency}, \textit{self-refinement}, \textit{multi-agent debate}, and \textit{mixture-of-agents}, to study their computational performance tradeoffs. 
We evaluate methods on two reasoning benchmarks (MMLU-Pro, BBH) and include extensive parameter configurations (e.g., scaling the number of parallel predictions, agents, and debate rounds) across different model sizes. 
Across 34 configurations and over 100 evaluations, we compute the Pareto-optimal front to select methods that achieve the best accuracy with the lowest computational budget.
Notably, inference scaling improves accuracy by up to +7.1\% points over chain-of-thought at the highest evaluated budgets (20× the CoT compute budget) on MMLU-Pro. 
With an equal computing budget, debate and mixture-of-agents outperform self-consistency by 1.3\% and 2.7\% points, respectively. 
While self-consistency saturates earlier, multi-agent gains persist, particularly on more complicated tasks. 
We identify a simple multi-agent design guideline: mixture-of-agents is most efficient when the number of parallel generations exceeds the number of sequential aggregations.
\end{abstract}

\begin{figure}[t!]
  \includegraphics[width=0.5\textwidth]{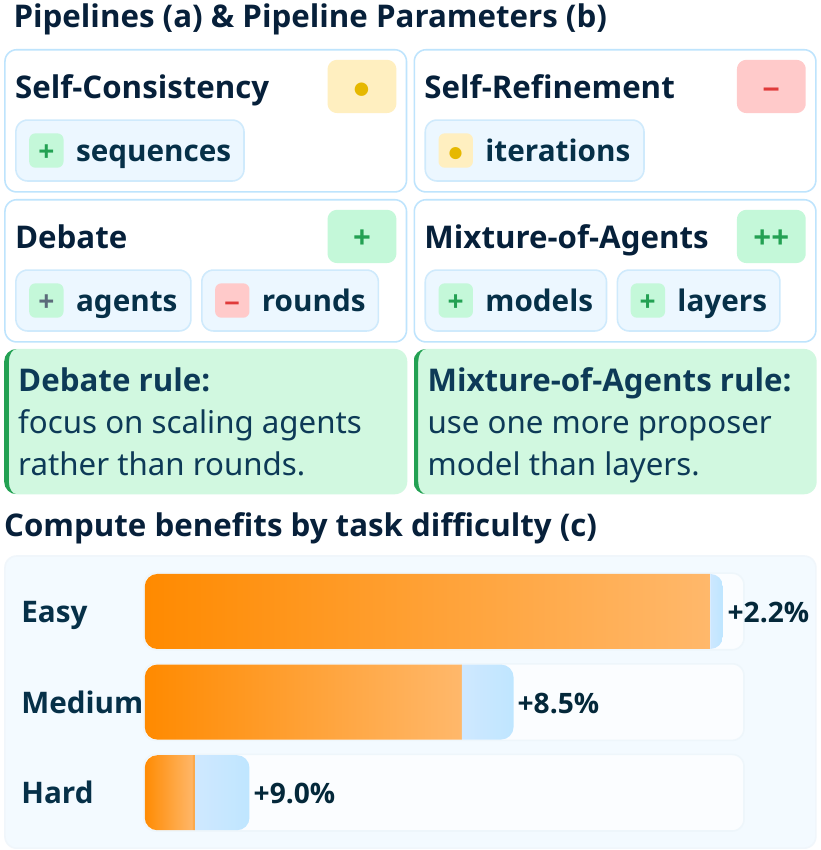}
    \caption{Test-time scaling efficiency ratings of: (a) Pipeline choice (self-consistency, self-refinement, debate, mixture-of-agents); (b) Pipeline parameters: green plus (recommended), red minus (discouraged), yellow circle (mixed results); (c) Chain-of-thought (orange) compared to the average performance gain across all tested pipeline configurations when using 15–20× the chain-of-thought compute budget (blue), across tasks of varying difficulty for MMLU-Pro.} 
\end{figure}

\section{Introduction}
The energy and computational demands of large language models (LLMs) are growing rapidly \cite{jinEnergyCostReasoning2025}. 
With increasing real-world applications, such as complementing search results at scale, efficiency becomes a core concern.
Many works focus on improving inference model performance without training, but use much larger inference compute budgets than others.
This makes comparing methods difficult as they can use orders of magnitude more compute than others.
The main goal of this study is to systematically analyze how to achieve the maximum performance gain within a given compute budget, as well as how to achieve a given performance level with minimal compute usage.

Inference-time strategies such as self-consistency \cite{wangSelfConsistencyImprovesChain2023}, self-refinement \cite{madaanSelfRefineIterativeRefinement2023}, debate \cite{duImprovingFactualityReasoning2023}, and mixture-of-agents (MoA) \cite{wangMixtureofAgentsEnhancesLarge2024} differ in their approach to scale inference compute, for example, through additional sampling, iterative refinement, or multi-agent orchestration. 
They also differ structurally \cite{zhangSurveyTestTimeScaling2025}: single-agent methods scale generations either in parallel with different samples (e.g., self-consistency) or sequentially with longer reasoning chains (e.g., self-refinement). 
Multi-agent systems such as MoA and debate use multiple interacting agents and scale along both axes simultaneously \cite{zhangOptimizingSequentialMultiStep2025, duImprovingFactualityReasoning2023, wangMixtureofAgentsEnhancesLarge2024}, thereby enabling interactions among different reasoning paths.

Studies have shown that scaling test-time compute can be more energy-efficient \cite{jinEnergyCostReasoning2025} and efficient with respect to the number of floating point operations (FLOPs) \cite{wuInferenceScalingLaws2024, snellScalingLLMTestTime2024} than increasing the model size alone.
Separately, work on multi-agent systems demonstrated performance gains over baselines such as chain-of-thought (CoT) \cite{duImprovingFactualityReasoning2023} or larger models \cite{wangMixtureofAgentsEnhancesLarge2024}. 
Despite rapid progress, three gaps remain: (i) Methods are frequently studied in isolation regarding their performance, but are not compared under matched compute budgets; (ii) Efficiency is usually measured with respect to FLOPs, rarely accounting for memory-transfer costs that affect processing time; (iii) Studies lack practical recommendations on method choice under constrained compute budgets (e.g., whether to use a single-agent method or a multi-agent method and how to parameterize it).

We address these gaps by analyzing four inference paradigms under matched compute budgets: self-consistency, self-refinement, debate, and MoA. To adjust compute budgets, we use three levers of test-time scaling: pipeline choice, pipeline parameters (agents, rounds, proposer models, layers, sequences, refinement iterations), and model size.
In addition to FLOPs, we also consider memory-transfer costs to measure compute times that are a closer approximation to practical applications on accelerators.
Finally, we provide recommendations for practitioners using multi-agent systems.

Our results show clear trends across benchmarks (MMLU-Pro, BBH) and model sizes (70B, 8B). 
Multi-agent pipelines are more efficient than single-agent baselines when scaled Pareto-optimally, with MoA dominating the Pareto-front. On MMLU-Pro, MoA improves the 70B-model accuracy from 64.3\% (CoT) to 71.4\% (+7.1 pp), outperforming self-consistency (68.7\%) and debate (70.0\%) by +2.7 and +1.4 percentage points under comparable inference budgets.
Pareto-optimal MoA efficiency is predominantly achieved when the number of parallel model generations exceeds the number of sequential aggregation steps by one. 
Larger models can be better and also more efficient than scaling up test-time compute with smaller models.
The greatest inference-time compute gains were observed on harder tasks (+9~pp for 15–20x CoT budget), motivating an adaptive routing approach that allocates more test-time scaling to harder tasks and less to easier tasks.
In general, we recommend a simple multi-agent design guideline: mixture-of-agents is one of the most robust and best-performing methods when the number of parallel generations exceeds the number of sequential aggregations.
\\

\noindent \textbf{\normalsize Key Contributions:}
\begin{itemize}[leftmargin=!, label={}, itemsep=0.3em, parsep=0pt]
    \item [\S\ref{sec:Method}\ \ \ ]
    We conduct a systematic study across four scaling methods (\textit{self-consistency}, \textit{self-refinement}, \textit{multi-agent debate}, \textit{mixture-of-agents}) and their parameterization.
    \item[\S\ref{Multi-Agent or Single-Agent}]%
    We demonstrate that multi-agent systems are more efficient than single-agent methods when scaled optimally. We find that mixture-of-agents consistently outperforms debate, self-consistency, and self-refinement.
    \item[\S\ref{How to scale multi-agent systems sequentially and in parallel}] We explore scaling behaviors of multi-agent systems and observe that configuring mixture-of-agents with one model more than layers predominantly yields Pareto-optimal performance, providing a practical design guideline.
    \item[\S\ref{How to choose the model size}] We show that larger models without test-time compute can be more compute-efficient than smaller models with heavily scaled test-time compute.
\end{itemize}

\section{Related Work} 
\label{sec:related-work}
Scaling laws for LLMs have historically focused on model parameter counts, training compute, and dataset size, with empirical studies showing predictable performance improvements as these factors increase \cite{kaplanScalingLawsNeural2020}. 
Recent research has shifted its focus toward optimizing test-time compute, as it enables improvements without requiring model retraining. 
As part of this development, various inference-time strategies have been introduced. 
Self-consistency \cite{wangSelfConsistencyImprovesChain2023} enhances reasoning abilities by sampling multiple CoT paths \cite{weiChainofThoughtPromptingElicits2023} and voting for the most consistent outcome. 
Self-refinement \cite{madaanSelfRefineIterativeRefinement2023} begins with an initial LLM output and then uses an iterative process of providing itself with feedback and refining its previous solution based on that feedback. 
Using multiple LLMs, MoA \cite{wangMixtureofAgentsEnhancesLarge2024} distributes the reasoning across different layers, with each layer having multiple proposer models for the first layers and an aggregator model for the final layer. 
Meanwhile, multi-agent debate \cite{duImprovingFactualityReasoning2023} uses multiple agents that iteratively refine their responses over a certain number of rounds using the responses of other agents as additional advice.

In addition to demonstrating improvements in accuracy, several studies have examined the efficiency of such strategies. \citet{jinEnergyCostReasoning2025} showed that scaling test-time compute can be more energy-efficient than scaling model parameters.
\citet{wuInferenceScalingLaws2024} and \citet{snellScalingLLMTestTime2024} demonstrated that small models with scaling strategies such as majority voting, best-of-n, and weighted voting can outperform larger models when using the same amount of FLOPs. However, \citet{sadhukhanKineticsRethinkingTestTime2025} observed that FLOP-centered estimates overstate the effectiveness of small models with test-time compute.
While most studies focus on FLOPs as the primary measure of compute, they neglect memory transfer, which also markedly influences runtime \cite{lindenliTransformerInferenceFirst2023}. 

Another limitation of previous work lies in the evaluation of multi-agent systems. The initial studies introducing debate \cite{duImprovingFactualityReasoning2023} and MoA \cite{wangMixtureofAgentsEnhancesLarge2024} evaluate these systems against single-model baselines.
MoA highlights superior efficiency compared to a single model call to a larger model, while debate demonstrates superior accuracy compared to CoT prompting with the same model. 
However, there has been little analysis that contrasts the performance of these multi-agent systems against single-agent approaches beyond simple CoT (e.g., self-consistency, self-refinement) and against other multi-agent systems under equal compute budgets. 

\citet{huangLargeLanguageModels2024} compare debate with self-consistency on the GMS8K benchmark \cite{cobbeTrainingVerifiersSolve2021} for an equal number of generations. 
They use three agents and two rounds in debate as in the initial debate paper \cite{duImprovingFactualityReasoning2023} and observe that debate does not outperform self-consistency under a similar compute budget.
However, their study does not compare debate with self-consistency under Pareto-optimal scaling of the number of agents and rounds. Additionally, the GMS8K benchmark is only limited to grade-school math \cite{cobbeTrainingVerifiersSolve2021}, whereas more diverse benchmarks such as MMLU-Pro \cite{wangMMLUProMoreRobust2024} or BBH \cite{suzgunChallengingBIGBenchTasks2022} are more robust to evaluate general reasoning performance.

In this work, we address these gaps by comparing the efficiency of single-agent pipelines containing self-consistency (parallel scaling) and self-refinement (sequential scaling) to multi-agent pipelines containing debate (parallel and sequential scaling with memory) and MoA (parallel and sequential scaling without memory). 
Our pipeline comparisons are based on a broad range of pipeline configurations (varying number of debate agents and rounds, MoA models and layers, refinement iterations, and self-consistency sequences). 
We identify compute-efficient pipeline and parameter choices and summarize them in practical rules of thumb for efficient scaling. 
Method costs are estimated by runtime, accounting for both FLOPs and memory transfer, and we evaluate all four pipelines with their varied parameters on the benchmarks MMLU-Pro and BBH that span a wide range of sub-tasks.

\section{Method} 
\label{sec:Method}

To explore how inference-time strategies influence reasoning efficiency, we vary compute along the levers pipeline choice (results in \Cref{Multi-Agent or Single-Agent}), pipeline parameters (results in \Cref{How to scale multi-agent systems sequentially and in parallel}), and model size (results in \Cref{How to choose the model size,MoA model sizes}), while assessing compute–accuracy trade-offs using Pareto-fronts.
For each generation, we assign the model the role of a reasoning expert that solves problems by thinking step by step via prompting. 
To extract multiple-choice answers at the end of each pipeline, we prompt the model to state its final answer choice (e.g., A/B/\dots), and select the completion option with the highest log-likelihood as the final answer. 
Prompt details for generations inside each pipeline are provided in \Cref{sec:prompts}.

Self-consistency samples multiple CoT paths and votes for the most consistent outcome. 
Self-refinement generates an initial solution and then iteratively improves it using its own feedback.
Debate consists of multiple agents that iteratively improve their responses over a fixed number of rounds, with each agent providing one answer per round.\footnote{We use the majority-vote variant, as voting-based debate scenarios outperform other decision protocols on reasoning tasks \cite{kaesbergVotingConsensusDecisionMaking2025, beckerMALLMMultiAgentLarge2025}.}
Lastly, MoA distributes reasoning across different layers, each of which has multiple proposer models for the first layers and an aggregator model for the last layer. 
Each proposer model generates an initial solution in the first layer and aggregates and synthesizes the responses from the previous layer in the layers $2, \dots,$ $l$$-$$1$. 
The aggregator model in the last layer aggregates the proposed solutions once more.\footnote{To mitigate model-specific biases, we adopt the single-proposer setting \cite{wangMixtureofAgentsEnhancesLarge2024}, where only a single LLM is used for all proposer models.}

\begin{figure*}[t!]
  \centering
  \includegraphics[width=0.77\textwidth]{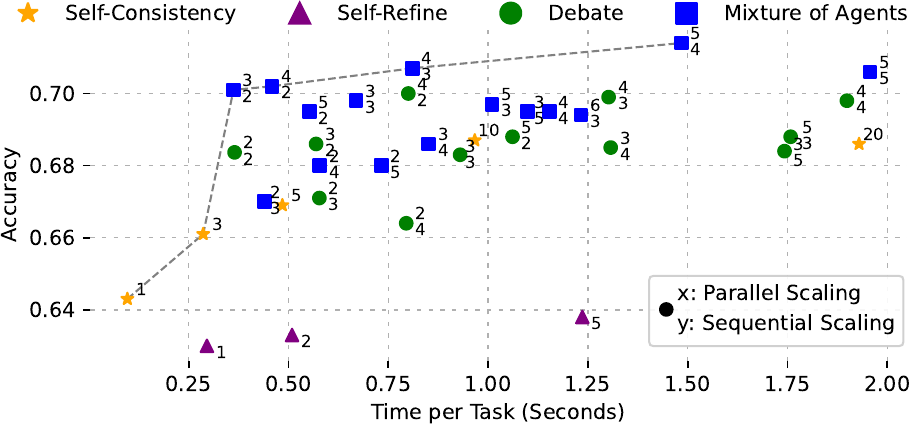}
  \caption{Accuracy (y-axis) and compute cost in time per task (x-axis) for multi-agent debate, MoA, self-consistency, and self-refinement on MMLU-Pro. Numbers to the lower-right of a point show the degree of parallel scaling and those to the upper-right the degree of sequential scaling. Self-consistency with one sequence is equivalent to CoT. The gray dotted Pareto-front shows the most efficient configuration per compute budget.
  }
  \label{fig:main plot mmlu pro 70b}
\end{figure*}

We vary pipeline parameters to scale test-time compute both in parallel (i.e., more independent attempts) and sequentially (i.e., greater depth per attempt).
Parallel variations occur in self-consistency (number of samples), debate (number of agents), and MoA (number of proposer models). Sequential variations occur in self-refinement (number of refinement steps), debate (number of debate rounds), and MoA (number of layers).
To examine the role of model size, we rerun the experiments shown in \Cref{fig:main plot mmlu pro 70b} by replacing a 70B-parameter model with an 8B-parameter model. We use models of the same family to separate the effects of inference-time scaling from those of model capacity.

To evaluate the compute-performance efficiency, we measure accuracy as the fraction of correctly answered questions and theoretically estimate the required computation time. 
Generation time is primarily influenced by two factors: the time required to perform operations and the time needed to transfer model weights from GPU memory to the compute cores. \Cref{tab:compute-estimation} in \Cref{sec:compute-estimation} presents the equations we use to estimate FLOPs, memory transfers, and execution time.

\section{Experiments}
\label{Experiments}
We conduct the primary experiments on the MMLU-Pro benchmark \cite{wangMMLUProMoreRobust2024}, which covers a wide range of topics. 
Our study verifies the robustness of selected results on the BBH benchmark \cite{suzgunChallengingBIGBenchTasks2022}. 
Since multi-agent systems are computationally expensive to run, we randomly shuffle tasks and draw a sample of 1000 questions. 
We provide the resulting confidence intervals for the accuracy in the limitations section. 
The main experiments employ Llama 3.1 70B, and we use Llama 3.1 8B when explicitly mentioned\footnote{\Cref{sec:compute-usage} provides model and quantization details.} \cite{grattafioriLlama3Herd2024}.

All generations are zero-shot with a temperature of 0.7 and a top-p of 0.95. 
After each pipeline, we select the answer candidate with the highest log-likelihood following the prompt \textit{``Final answer of choices \{choices\}:``} as the final answer. 
To ensure fair comparisons, we keep as many parameters constant as possible and vary only those relevant to each experiment. We release the code for our experiments publicly\footnote{\url{https://github.com/Multi-Agent-LLMs/lm-evaluation-harness}}.

\subsection{Are multi-agent systems more compute-\newline efficient than single-agent methods?}
\label{Multi-Agent or Single-Agent}

Multi-agent systems can outperform single-agent baselines \cite{wangMixtureofAgentsEnhancesLarge2024, duImprovingFactualityReasoning2023}, but they typically require more compute.
This raises the question of whether they improve mainly because of compute scaling, or if they are also more efficient than scaling up single-agent systems.
To examine this, we evaluate the multi-agent systems debate and MoA for varying configurations of agents and layers (debate) as well as models and layers (MoA), and compare both performance and compute to the single-agent self-consistency baseline and self-refinement.

\Cref{fig:main plot mmlu pro 70b} shows the compute-accuracy tradeoffs on the MMLU-Pro benchmark. 
MoA configurations consistently lie closer to the Pareto-front than self-consistency, thereby delivering higher compute-accuracy efficiency. 
Similarly, debate is more efficient than self-consistency when using more agents than rounds. 
Across the tested range of up to \textasciitilde20×CoT compute, the highest accuracies are 68.7\% for self-consistency (10 iterations), 70.0\% for debate (4 agents, 2 layers), and 71.4\% for MoA (5 models, 4 layers). 
Compared to the CoT baseline at 64.3\%, this corresponds to gains of +4.4\%, +5.7\%, and +7.1\% points, respectively. 
Relative to self-consistency under comparable compute budgets, debate and MoA gain +1.3 and +2.7 percentage points, solely by using a different scaling paradigm. Self-refinement consistently yields lower accuracy than the CoT baseline despite requiring more compute. 
Both self-consistency and debate exhibit saturation at around 10 sequences and 4 agents, respectively, whereas MoA continues to improve nearly linearly up to 5 models and 4 layers.

Our results suggest that multi‑agent systems not only achieve better performance but are also more efficient than single‑agent setups.
The observation that MoA scales further than debate may be explained by its architecture.
Considering a fixed number of proposer models, adding layers does not increase the context length per model \cite{wangMixtureofAgentsEnhancesLarge2024}.
However, in debate, increasing the number of rounds with a fixed number of agents expands the context, as agents retain prior exchanges in memory \cite{duImprovingFactualityReasoning2023}, thereby increasing computational overhead. 
It is surprising to see that self-refinement performed worse than the CoT baseline, but it aligns with the finding of \citet{huangLargeLanguageModels2024}.
We recommend MoA over the use of debate, self-consistency, and self-refinement. However, the efficiency depends not only on the pipeline itself but also on how the parameters are set, which we address in \Cref{How to scale multi-agent systems sequentially and in parallel}.

\subsection{Do difficult tasks benefit more from test-time scaling than easy ones?}
\label{How does scaling behave with respect to task difficulty?}

In the previous section, we observed a trend in which increasing test-time compute improves task performance. 
However, we want to examine whether this holds across all difficulty levels. 
While some tasks benefit from increased compute usage, others may not require it or could even suffer from an overthinking phenomenon \cite{ghosalDoesThinkingMore2025,chenNOTThinkThat2024}.
To investigate this, we first estimate task difficulties on MMLU-Pro using CoT solve rates. 
We let the model answer each of the 1,000 questions 20 times with CoT prompting and assign each task to one of three bins: easy (solve rate >75\%), medium (25–75\%), and hard (<25\%). 
Based on this grouping, we analyze how compute scaling impacts performance across difficulties.

\begin{table}[t]
\centering
\begin{tabular}{l|c:c:c}
\textbf{Compute} & \textbf{Easy} & \textbf{Medium} & \textbf{Hard} \\
\hline
CoT & 94.4 & 53.0 & 8.4 \\
1-5 $\times$ CoT & 95.6 & 58.6 & 13.6 \\
5-10 $\times$ CoT & 95.4 & 60.4 & 14.7 \\
10-15 $\times$ CoT & 96.0 & 62.1 & 14.2 \\
15-20 $\times$ CoT & 96.6 & 61.5 & 17.4 \\
\hdashline
Gain ($\Delta$) & +2.2 & +8.5 & +9.0
\end{tabular}
\caption{Accuracy on easy, medium and hard tasks for different compute bins relative to the CoT compute. ``Easy'': more than 75\% solve rate, ``Medium'': 25\%-75\% and ``Hard'': less than 25\% CoT solve rate. 
$\Delta$ is the accuracy gain between 15–20 × CoT and CoT compute.}
\label{tab:task difficulty vs compute budget}
\end{table}

\Cref{tab:task difficulty vs compute budget} shows the accuracy gains for easy, medium, and hard tasks.
Easy tasks have a 94.4\% CoT accuracy, which improves by 2.2\% points when compute is increased to an equivalent of 15–20 times the CoT compute. 
Medium-difficulty tasks start with a CoT accuracy of 53.0\%, which improves by 8.5\% points, and hard tasks have an 8.4\% CoT accuracy, which gains 9.0\% points as test-time compute increases.

The solve rates improve across all difficulty levels, with the largest absolute gain on hard tasks (+9.0\%), followed by medium-difficulty tasks (+8.5\%). 
For hard tasks, this corresponds to more than doubling the accuracy relative to the CoT baseline.
This indicates that additional test-time compute is particularly valuable for more challenging problems.
At the same time, easy tasks exhibited only marginal gains, suggesting that a routing mechanism that assigns them to less test-time scaling could preserve accuracy while avoiding unnecessary compute costs. 
This idea aligns with adaptive frameworks such as BEST-Route \cite{dingBESTRouteAdaptiveLLM2025}, which allocate compute based on task difficulty to optimize efficiency.

\subsection{What is the optimal scaling ratio that maximizes debate and MoA efficiency?}

\label{How to scale multi-agent systems sequentially and in parallel}

Although the general trend of increasing performance with increasing test-time compute holds, there are noticeable differences in efficiency across configurations of the same multi-agent setup. 
Scaling efficiently allows for achieving better results given the same amount of compute or the same results using less compute and is therefore advisable for large-scale deployment in real-world applications.
Using our evaluations on the MMLU-Pro benchmark, we visualize how parallel and sequential scaling affect accuracy and compute usage in the multi-agent systems debate and MoA. 

\begin{figure}[t]
  \includegraphics[width=\linewidth]{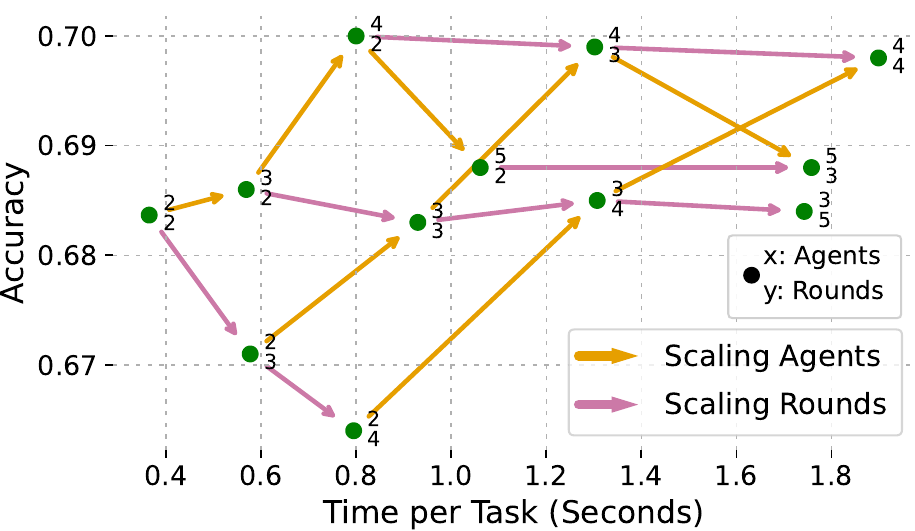}
  \caption{Effect of scaling agents (orange) and rounds (magenta) in multi-agent debate on the MMLU-Pro benchmark. Superscript values indicate the number of agents, subscript values the number of rounds.}
  \label{fig:Debate Scaling}
\end{figure}

\Cref{fig:Debate Scaling} shows the effect of parallel scaling of agents (orange) and sequential scaling of rounds (magenta) in the debate mechanism. 
Accuracy improves when scaling agents up to four, and decreases with more agents. 
Scaling rounds degrades performance with two agents and has no noticeable effect with more agents.

Our result indicates that scaling rounds can negatively affect performance, aligning with prior work and resembling limitations of other sequential scaling approaches, such as self-refinement \cite{kaesbergVotingConsensusDecisionMaking2025,beckerStayFocusedProblem2025}.
Nevertheless, the effect is not consistent.
When testing the robustness of the result on BBH and with an 8B-parameter model, respectively (not shown here but in \Cref{fig:BBH plot,fig:8B plot} of \Cref{Different Evaluation Setups}), increasing the number of debate rounds leads to improvements in accuracy in some cases. 
However, the highest debate accuracy is consistently achieved with only two rounds, suggesting a practical recommendation, i.e., prioritize scaling agents over rounds, as parallel scaling yields more reliable gains.

\begin{figure}[t]
  \includegraphics[width=\linewidth]{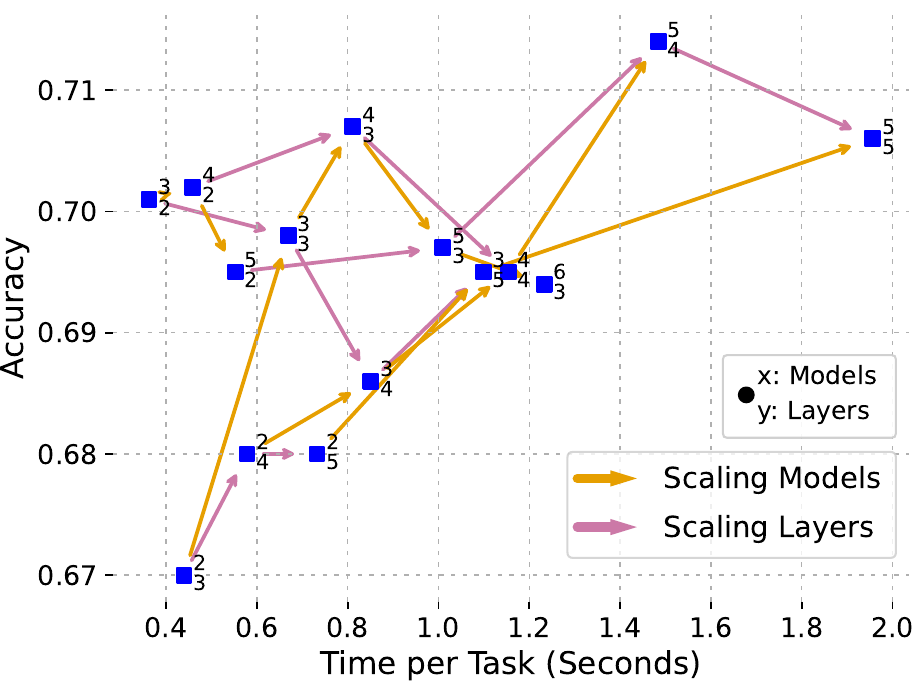}
  \caption{Effect of scaling proposer models (orange) and layers (magenta) in MoA on the MMLU-Pro benchmark. Superscript values indicate the number of proposer models; subscripts the number of layers.}
  \label{fig:MoA Scaling}
\end{figure}

\Cref{fig:MoA Scaling} shows the effect of parallel scaling of proposer models (orange) and sequential scaling of layers (magenta) for MoA. 
Accuracy improves as models or layers are scaled toward the point where the number of models exceeds the number of layers by one, with Pareto-optimal performance observed at this specific ratio (3 models with 2 layers, 4 models with 3 layers, and 5 models with 4 layers). Beyond this point, adding more models or layers tends to reduce accuracy.

This result shows that both parallel and sequential scaling are beneficial in the MoA system. We verify across additional evaluations on the BBH benchmark and with an 8B model that it is beneficial to use one model more than layers (see \Cref{fig:BBH plot,fig:8B plot} in \Cref{Different Evaluation Setups}). In both cases, two of the three points following the rule lie on the Pareto-optimal front, confirming its ability to achieve Pareto-optimal efficiency. 
Sequential scaling is less detrimental than in debate or self-refinement, likely because the context is not accumulated across layers.

\subsection{Are heavily scaled small models more compute-efficient than larger models?}
\label{How to choose the model size}

Larger models typically offer stronger reasoning capabilities but incur higher compute and memory requirements, leading to increased inference latency due to the heavier load. 
While multi-agent scaling can improve performance, it may exhibit diminishing returns. 
This raises the key question: given equal compute budgets, is it more efficient to scale multi-agent setups further or should you instead transition to a larger model?
To explore potential transition points between smaller and larger models, we compare multi-agent setups scaled with Llama 3.1 8B and less scaled setups using Llama 3.1 70B.

Across all pipeline setups (debate, MoA, self-consistency, and self-refinement) and their varied parameters, the smaller model achieves accuracies below 53\% on MMLU-Pro. 
This is lower than the CoT baseline of the 70B-parameter model (64.3\%), although the most scaled-up 8B configurations require more compute. 
With the same compute budget as the 70B CoT baseline, the best 8B configuration is approximately 13\% less accurate than the 70B CoT pipeline. 
Details about accuracy and compute across tested 8B configurations are not shown here but in \Cref{fig:8B plot} in \Cref{Different Evaluation Setups}.

The observed accuracy difference shows that model choice and size can have a larger impact on efficiency than test-time compute scaling alone. In our experiments, given sufficient computational resources to run the 70B-parameter model with CoT, it would be advantageous to use the larger model. 
In contrast, prior work has reported that smaller models enhanced with test-time compute techniques, such as majority vote, can outperform substantially larger models relying solely on scale \cite{jinEnergyCostReasoning2025}. 
The impact of test-time compute scaling compared to using a larger model likely depends on model strength and quantization effects. In our study, we used models from the same family to minimize confounding factors, though the quantization might have impacted the 8B model more strongly than the 70B model.

\subsection{Is MoA more compute-efficient with mixed-size or uniform-size models?}
\label{MoA model sizes}
Since the MoA system outperformed the other scaling approaches in our experiments (see \Cref{Multi-Agent or Single-Agent}), we further investigate how its performance depends on the allocation of model sizes within the system. 
The MoA pipeline consists of multiple proposer models in layers $1, \dots,$ $l$$-$$1$ that aggregate and synthesize candidate solutions from the previous layer and an aggregator model in the last layer $l$ \cite{wangMixtureofAgentsEnhancesLarge2024}. 
In all our previous experiments, we used the same model size for both the proposers and the aggregator. 
However, these roles may influence efficiency differently. 
This raises the question of whether it is more computationally efficient to use a larger model as the aggregator and smaller ones as proposers, or vice versa.
We reevaluated the MoA configuration with five models and four layers by substituting the 70B-parameter model with the smaller 8B model in the proposer or aggregator role, while keeping the rest of the setup fixed.

\Cref{tab:MoA proposer aggregator model sizes} summarizes the results.
Using 70B-parameter models for both proposer and aggregator models achieves the highest MMLU-Pro accuracy of 71.4\%.
Replacing the aggregator with an 8B model while keeping the proposers at 70B leads to only a minor drop to 69.6\%. 
In contrast, reducing the proposers to 8B models while keeping a 70B aggregator decreases accuracy to 52.9\%. 
The lowest performance occurs when both the proposer and aggregator models are reduced to 8B parameters, yielding 51.2\% accuracy.

\begin{table}[t]
\centering
\begin{tabular}{l|cc}
   MoA Accuracy & \textbf{8B Prop.} & \textbf{70B Prop.} \\ \midrule
\textbf{8B Aggr.} & 51.2 & 69.6 \\
\textbf{70B Aggr.}  &  52.9 & 71.4 \\
\end{tabular}
\caption{MoA accuracy (five models, four layers) with various proposer and aggregator model sizes on the MMLU-Pro benchmark.}
\label{tab:MoA proposer aggregator model sizes}
\vspace{-0.5em}
\end{table}

The results show that the proposer size has a greater influence on accuracy than the aggregator size. 
When proposers are strong, the aggregator operates over higher-quality evidence and can remain relatively small without large losses. 
When proposers are weak, the downstream information is already degraded; a single strong aggregator cannot reliably recover quality that was never generated. 
In terms of compute, the aggregator size influences the total cost less than the proposer sizes, since it only aggregates once in the final layer.
In contrast, each of the five proposers generates once in the three previous layers, yielding 15 generations. 
Thus, the aggregator contributes only $1/16$ of all generations in this example.
In sum, proposers are most important for quality, and the aggregator requires only a single generation, thereby influencing compute only marginally. These results suggest using a similar model size for proposer and aggregator models.%

In our previous experiments, all proposer models were either large or small. 
Since proposers are the main drivers of answer quality but also dominate compute costs, we further explore whether mixing smaller and larger proposer models can reduce compute costs without noticeably harming accuracy. 
Intuitively, smaller models can generate diverse candidate solutions at a lower cost, while larger models may better identify flaws or refine proposals.
We reevaluate the MoA configuration with five models and four layers, this time varying the ratio of smaller (8B) to larger (70B) proposer models while keeping the aggregator fixed at 70B parameters.

\begin{figure}[t]
  \vspace{-0.5em}
  \includegraphics[width=\linewidth]{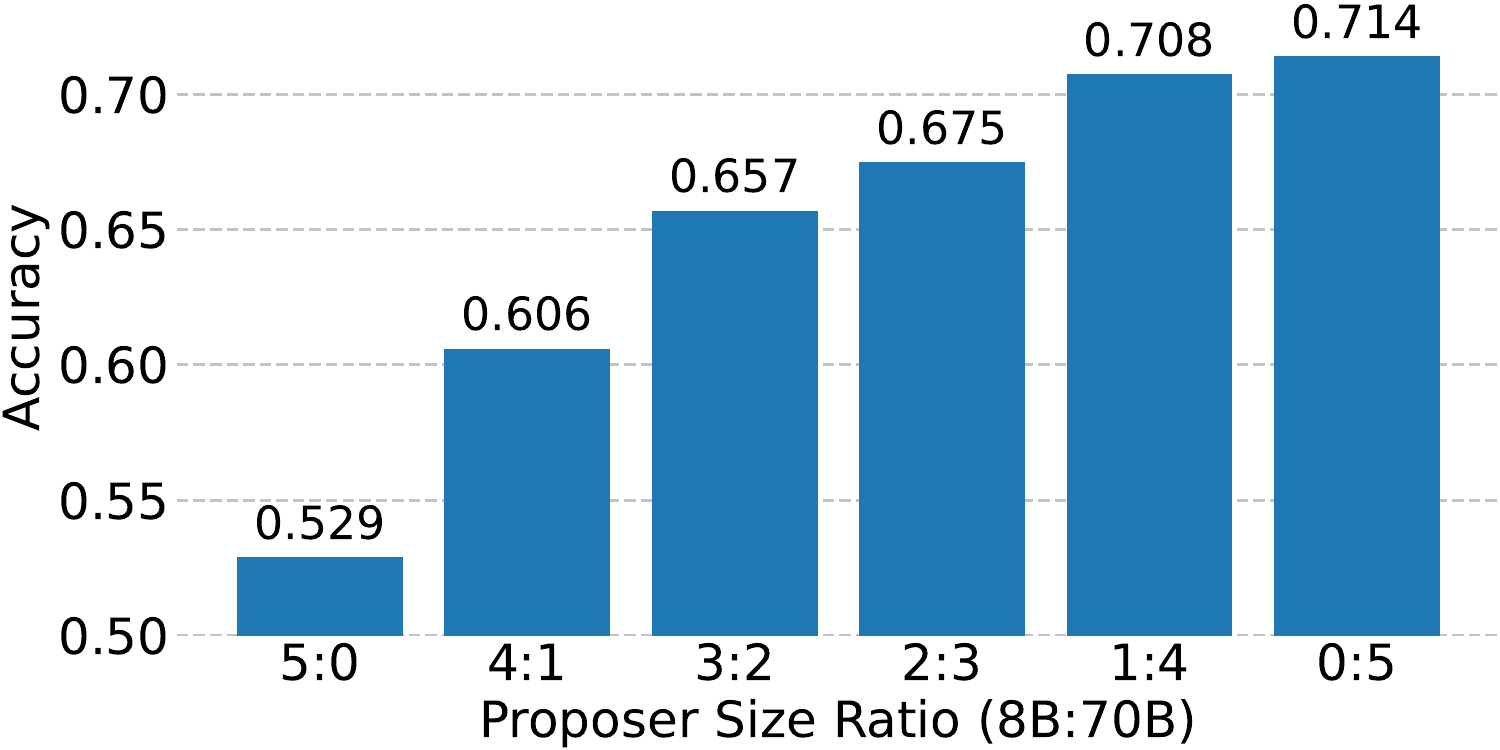}
  \caption{MoA accuracy for different ratios of 8B and 70B proposer model sizes in the configuration of five models with four layers (8B:70B ratio) on MMLU-Pro.}
  \label{fig:proposerSizeRatios}
  \vspace{-0.8em}
\end{figure}

\Cref{fig:proposerSizeRatios} shows that the accuracy increases monotonically with the proportion of larger 70B models. 
Configurations with one small 8B proposer out of five show a minor reduction in performance compared to the all-large setup. 
As the number of smaller proposer models grows, performance declines more substantially, reaching the lowest accuracy when all proposers have 8B parameters.

While using a single small 8B proposer has little impact on accuracy, it yields only minimal compute savings because it accounts for only a small fraction of proposer generations. 
In some cases, adding small proposers can even reduce performance over entirely leaving them out: the all-large (70B) MoA with three proposers and two layers achieves 70.1\% accuracy on MMLU-Pro (see \Cref{fig:main plot mmlu pro 70b}), while a mixed-size MoA with five models (three 70B and two 8B models) reaches only 67.5\%, despite both setups following the Pareto-optimal scaling rule from \Cref{How to scale multi-agent systems sequentially and in parallel} with one model more than layers. 
Overall, these results suggest that using proposer models of the same size is generally more effective than mixing models of different sizes.

\vspace{-0.8em}
\section{Conclusion}
\vspace{-0.5em}
We analyzed the test-time scaling of LLMs via single-agent and multi-agent setups on a diverse range of reasoning tasks using MMLU-Pro and BBH, and assessed their compute-accuracy trade-offs. 
We controlled for compute in terms of runtime that accounts for both FLOP and memory transfer times. 
Our experiments systematically explored three levers of test-time scaling: (1) the choice of inference pipeline, (2) the adjustment of pipeline parameters, and (3) the choice of model size.

Across the pipelines self-consistency, self-refinement, debate, and MoA, we found that multi-agent systems consistently outperform single-agent approaches, with MoA dominating the Pareto-front across benchmarks. 
Under comparable compute budgets, debate yielded a 1.3\% increase in accuracy, and MoA a 2.7\% increase over self-consistency, driven solely by their different scaling paradigms. 
Analyzing scaling in multi-agent pipelines revealed clear design principles: in debate, where memory accumulates over several rounds, efficiency gains primarily arise from scaling the number of agents rather than the number of rounds. 
In MoA, Pareto-optimal efficiency is often achieved when the number of proposer models exceeds the number of layers by one. 
This paradigm increased accuracy by 7.1\% points at \textasciitilde20× the CoT budget; but if compute is tight, a \textasciitilde5×CoT budget already captures most of that gain. 
When MoA is not feasible ($\leq$ 3×CoT budget), self-consistency is the most viable option.

We further showed that model size can exert a stronger influence on efficiency than test-time scaling alone. Larger models not only deliver higher performance but can also be more compute-efficient than heavily scaled smaller models. 
In addition, our analysis across task difficulty levels demonstrated that additional test-time compute is most beneficial for harder tasks, whereas easy tasks exhibit diminishing returns. 
This motivates adaptive routing approaches that dynamically allocate compute based on task difficulty, optimizing both efficiency and accuracy.

Finally, our study underlines the value of evaluating inference strategies through efficiency metrics, as benchmark accuracy alone can often be achieved by allocating additional compute.
This perspective motivates future work to emphasize efficiency under deployment-relevant cost models that account for both FLOPs and memory transfer.

\section*{Limitations}
 Since multi-agent systems are computationally expensive to run, we used a sample size of n=1000 instead of the full benchmark. 
 This introduces some variance. 
 The 95\% binomial confidence interval for the true accuracy of each data point can be calculated using the measured accuracy $\hat{p}$:
\[
\hat{p} \pm 1.96 \cdot \sqrt{\frac{\hat{p}(1 - \hat{p})}{n}}
\]

For measured accuracies between 0.63 and 0.72, the uncertainty ranges from 0.028 to 0.03.
While individual configurations may have overlapping 95\% confidence intervals, clear patterns emerge when examining the results as a whole, as patterns persist for similar configurations (e.g., other benchmarks, model size, near-the-same pipeline parameters).

Our compute costs were estimated using theoretical formulas that account for FLOPs and memory transfer. 
This provides a consistent basis for comparison; however, real-world runtime may vary due to framework overhead, batching strategies, or system-level constraints. 
These practical considerations also matter when interpreting our finding that larger models, such as Llama 3.1 with 70B parameters, can be more computationally efficient than scaling smaller models extensively. 
In deployment scenarios with limited GPU memory, such efficiency gains may not be realizable, making it important to explore trade-offs under constrained hardware conditions.
Beyond MMLU-Pro and BBH, reasoning benchmarks that target specific capabilities, such as spatial pathfinding \citep{kaesberg-etal-2025-sparc} or step-level agent evaluation in interactive environments \citep{kaesberg2026mindgapspatialreasoning}, remain unexplored and may exhibit different compute-accuracy trade-offs.

Finally, our study was limited to non-reasoning models. Future work could explore how reasoning models interact with inference-time scaling strategies, for example, whether their stronger reasoning abilities change the compute-performance trade-offs and what patterns (e.g., reasoning strength, pipeline parameters) lead to the most efficient scaling of multi-agent systems with reasoning models.

\section*{Acknowledgments}
This work was partially supported by the Lower Saxony Ministry of Science and Culture and the VW Foundation. Many thanks to Niklas Bauer for his thoughtful discussions and feedback. This work used the Scientific Compute Cluster at GWDG, the joint data center of Max Planck Society for the Advancement of Science (MPG) and University of Göttingen. In part funded by the Deutsche Forschungsgemeinschaft (DFG, German Research Foundation) – 405797229.

\bibliography{latex/custom}

\appendix
\newpage
\section{Models \& Compute Usage}
\label{sec:compute-usage}

Our experiments used 4-bit quantized Llama 3.1 models (\textit{hugging-quants/Meta-Llama-3.1-70B-Instruct-GPTQ-INT4} with commit hash \textit{1b0ae7f\allowbreak 9d6da8\allowbreak b79f36fd\allowbreak c24912f9\allowbreak 50ecb2b6e91} and \textit{hugging-quants/Meta-Llama-3.1-8B-Instruct-GPTQ-INT4} with commit hash \textit{f184d2\allowbreak b18e2b7c\allowbreak 257af7\allowbreak 9aa0b\allowbreak c53ef3ee\allowbreak 505caa2}). The 8B model was used for experiments on MMLU-Pro and the 70B model was used for experiments on both MMLU-Pro and BBH. Models were hosted on a node of 8 NVIDIA A100 40GB PCIe GPUs.
The number of processed tokens and the resulting floating point operations (FLOPs) are provided in \Cref{tab:compute-usage1}.

\begin{table}[h]
\centering
\begin{tabular}{l|cc}
\makecell{Model} & \makecell{Tokens} & \makecell{FLOPs}\\
\hline
Llama 3.1 70B* & $1.56 * 10 ^ 9$ & $217.87 * 10 ^ {18}$\\
Llama 3.1 8B* & $1.37 * 10 ^ 9$ & $21.95 * 10 ^ {18}$\\
\hdashline
\textbf{Sum} & $2.93 * 10 ^ 9$ & $239.82 * 10 ^ {18}$\\
\end{tabular}
\caption{Compute usage (tokens, FLOPs) across all experiments for the 4-bit quantized Llama 3.1 models (\textit{hugging-quants/Meta-Llama-3.1-70B-Instruct-GPTQ-INT4} and \textit{hugging-quants/Meta-Llama-3.1-8B-Instruct-GPTQ-INT4}). FLOPs are calculated with the formulas from \Cref{tab:compute-estimation} of Appendix \ref{sec:compute-estimation}.}
\label{tab:compute-usage1}
\end{table}

\section{Compute Estimation Formulas}
\label{sec:compute-estimation}

Generation time is primarily determined by two factors: the time required to perform arithmetic operations and the time needed to transfer model weights from the GPU memory into the computation cores. In addition to these factors, generation can be divided into two stages: a prefill stage, in which the model processes all input tokens once to produce and cache key-value states, and a decode stage, in which the model generates output tokens one by one, reusing the cached states \cite{lindenliTransformerInferenceFirst2023}.

For each stage, we estimate both the compute time and the memory transfer time, following the formulation of \citet{lindenliTransformerInferenceFirst2023}, with the specific equations summarized in \Cref{tab:compute-estimation}. Since computation and memory transfer occur in parallel, the slower of the two determines the effective runtime. FLOPs are estimated by assuming that each parameter is used for one multiplication and one addition per token. Memory transfer is estimated based on how often the model weights must be loaded from the GPU memory into the computation cores, which happens once during each prefill stage and repeatedly during decoding. Both FLOPs and memory transfers are converted to times using the GPU’s peak compute throughput and memory bandwidth.

For all estimations, we assumed a batch size of 16 and used the Llama 3.1 models (70B and 8B parameters) with 4-bit quantization, corresponding to a model precision of 0.5 bytes per parameter. Furthermore, all estimations were made with 8 NVIDIA A100 40GB PCIe GPUs, each with a peak throughput of 1,248 tera-operations per second (for 4-bit computation) and a memory bandwidth of 1,555 GB/s \cite{nvidiacorporationNVIDIAA100Tensor2021}.

\begin{table}[thb]
\centering
\small
\renewcommand{\arraystretch}{1.15}
\setlength{\tabcolsep}{0pt}
\begin{tabular}{@{}l@{}}
\textbf{FLOPs} \\[0.25em]
$\text{FLOPs}_{\text{prefill}} = 2 \cdot P \cdot T_\text{in}$ \\
$\text{FLOPs}_{\text{decode}} = 2 \cdot P \cdot T_\text{out}$ \\
$\text{FLOPs}_{\text{total}} = \text{FLOPs}_{\text{prefill}} + \text{FLOPs}_{\text{decode}}$ \\[0.6em]

\textbf{Model Memory} \\[0.25em]
$\text{Mem}_\text{model} = P \cdot p$ \\[0.6em]
\textbf{Memory Transfer}\\ 
$\text{Mem}_{\text{prefill}} = \frac{G}{B} \cdot \text{Mem}_\text{model}$ \\[0.1em]
$\text{Mem}_{\text{decode}} = \frac{T_\text{out}}{B} \cdot \text{Mem}_\text{model}$ \\
$\text{Mem}_{\text{total}} = \text{Mem}_{\text{prefill}} + \text{Mem}_{\text{decode}}$ \\[0.6em]

\textbf{FLOP Time} \\[0.25em]
$\text{Time}_{\text{flop,prefill}} = \frac{\text{FLOPs}_{\text{prefill}}}{N_\text{GPU} \cdot F_\text{GPU}}$ \\[0.4em]
$\text{Time}_{\text{flop,decode}} = \frac{\text{FLOPs}_{\text{decode}}}{N_\text{GPU} \cdot F_\text{GPU}}$ \\[0.6em]

\textbf{Memory Transfer Time} \\[0.25em]
$\text{Time}_{\text{mem,prefill}} = \frac{\text{Mem}_{\text{prefill}}}{N_\text{GPU} \cdot M_\text{GPU}}$ \\[0.4em]
$\text{Time}_{\text{mem,decode}} = \frac{\text{Mem}_{\text{decode}}}{N_\text{GPU} \cdot M_\text{GPU}}$ \\[0.6em]

\textbf{Total Time} \\[0.25em]
$\text{Time}_{\text{prefill}} = \max\left(\text{Time}_{\text{flop,prefill}},\, \text{Time}_{\text{mem,prefill}}\right)$ \\
$\text{Time}_{\text{decode}} = \max\left(\text{Time}_{\text{flop,decode}},\, \text{Time}_{\text{mem,decode}}\right)$ \\
$\text{Time}_{\text{total}} = \text{Time}_{\text{prefill}} + \text{Time}_{\text{decode}}$
\end{tabular}
\caption{Formulas used for compute estimation. $P$: model parameters, $p$: model precision (bytes per parameter), $T_\text{in}$/$T_\text{out}$: number of input/output tokens, $G$: number of generations, $B$: batch size, $N_\text{GPU}$: number of GPUs, $F_\text{GPU}$: FLOPs/s per GPU, $M_\text{GPU}$: memory bandwidth per GPU --- build on formulas from \citet{lindenliTransformerInferenceFirst2023}.}
\label{tab:compute-estimation}
\end{table}

\section{Prompts}
\label{sec:prompts}
In the following, we list the prompts used in our evaluation. The model was instructed to adopt the role of an expert in reasoning and to solve given tasks step by step. For pipeline-specific prompts, we follow the original formulations from \citet{wangSelfConsistencyImprovesChain2023} (self-consistency), \citet{madaanSelfRefineIterativeRefinement2023} (self-refinement), \citet{duImprovingFactualityReasoning2023} (debate) and \citet{wangMixtureofAgentsEnhancesLarge2024} (MoA). To extract the final answer, the model was prompted to state its answer choice, and the option with the highest log-likelihood was selected as the final answer.

\begin{tcolorbox}[colback=yellow!5,colframe=gray,title=Self-consistency system prompt]
You are a reasoning expert. Solve the task provided by the user while thinking step by step. Make sure to state your answer at the end of the response.
\end{tcolorbox}
\begin{tcolorbox}[colback=purple!5,colframe=gray,title=Self-refinement system prompt]
You are a reasoning expert. Solve the tasks provided by the user while thinking step by step. Make sure to state your answer at the end of the response. When asked to solve a task, only solve the task whilst thinking step by step. When asked to provide feedback, only provide feedback without solving the task.
\end{tcolorbox}
\begin{tcolorbox}[colback=purple!5,colframe=gray,title=Self-refinement feedback prompt]
Provide actionable and specific feedback to your last response that suggests a clear improvement. Actionable means that the feedback should contain a concrete action that would likely improve the response. Specific means that the feedback should identify concrete phrases in the output to change. The suggested improvement should increase the chance of correctly solving the given task.
\end{tcolorbox}
\begin{tcolorbox}[colback=purple!5,colframe=gray,title=Self-refinement refinement prompt]
Use the feedback to produce a refined solution.
\end{tcolorbox}
\begin{tcolorbox}[colback=green!5,colframe=gray,title=Debate system prompt]
You are a reasoning expert. Solve the task provided by the user while thinking step by step. Make sure to state your answer at the end of the response.
\end{tcolorbox}
\begin{tcolorbox}[colback=green!5,colframe=gray,title=Debate debating prompt]
These are the solutions to the problem from other agents:\\

Agent 1's solution: ```\{agent\_1\_solution\}```\\

...\\

Agent n's solution: ```\{agent\_n\_solution\}```\\

Using the opinion of other agents as additional advice, can you give an updated response
\end{tcolorbox}
\begin{tcolorbox}[colback=blue!5,colframe=gray,title=MoA system prompt]
You are a reasoning expert. Solve the task provided by the user while thinking step by step. Make sure to state your answer at the end of the response.
\end{tcolorbox}
\begin{tcolorbox}[colback=blue!5,colframe=gray,title=MoA aggregation prompt]
You have been provided with a set of responses from various open-source models to the latest user query. Your task is to synthesize these responses into a single, high-quality response. It is crucial to critically evaluate the information provided in these responses, recognizing that some of it may be biased or incorrect. Your response should not simply replicate the given answers but should offer a refined, accurate, and comprehensive reply to the instruction. Ensure your response is well-structured, coherent, and adheres to the highest standards of accuracy and reliability.\\

Responses from models:\\
```\{response\_model\_1\}```\\

...\\

```\{response\_model\_n\}```
\end{tcolorbox}
\begin{tcolorbox}[colback=black!5,colframe=gray,title=Answer extraction prompt]
Final answer of choices \{choices\_str\}:
\end{tcolorbox}

\onecolumn
\section{Different Evaluation Setups}
\label{Different Evaluation Setups}
\begin{figure}[h]
  \centering
  \includegraphics[scale=0.75]{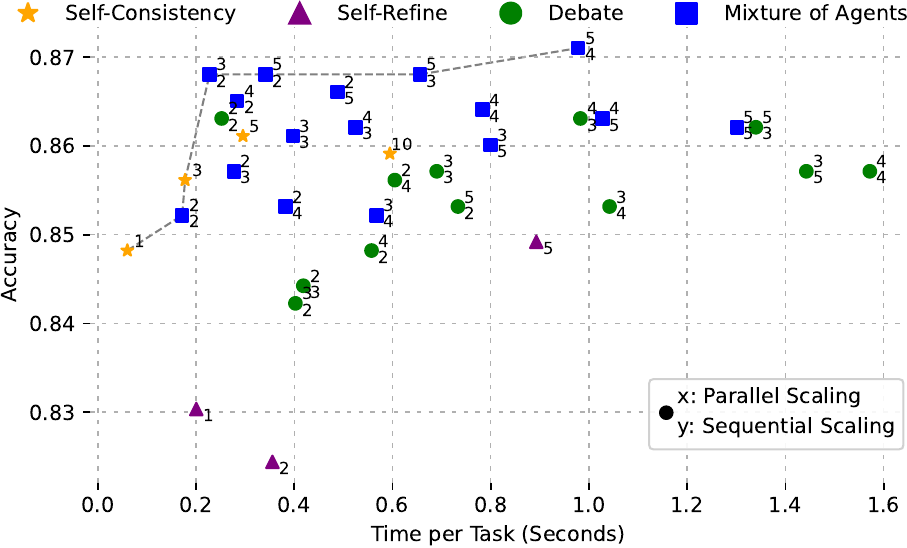}
  \caption{Accuracy (y-axis) and compute cost in time per task (x-axis) for multi-agent debate, MoA, self-consistency, and self-refinement on BBH \cite{suzgunChallengingBIGBenchTasks2022}. Numbers to the lower-right of a point show the degree of parallel scaling and those to the upper-right the degree of sequential scaling. Self-consistency with one sequence is equivalent to CoT. The gray dotted Pareto-front shows the most efficient configuration per compute budget.}
  \label{fig:BBH plot}
\end{figure}

\begin{figure}[h]
  \centering
  \includegraphics[scale=0.75]{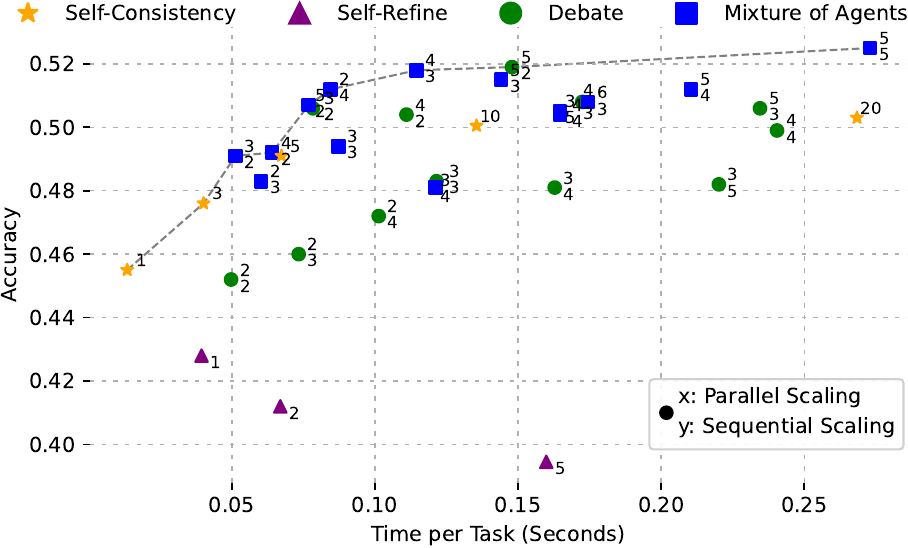}
  \caption{Accuracy (y-axis) and compute cost in time per task (x-axis) for multi-agent debate, MoA, self-consistency, and self-refinement on MMLU-Pro \cite{wangMMLUProMoreRobust2024} with the 8B-parameter model. Numbers to the lower-right of a point show the degree of parallel scaling and those to the upper-right the degree of sequential scaling. Self-consistency with one sequence is equivalent to CoT. The gray dotted Pareto-front shows the most efficient configuration per compute budget.}
  \label{fig:8B plot}
\end{figure}

To assess the robustness of our results, we reran the experiments from \Cref{fig:main plot mmlu pro 70b} with one differing factor. In \Cref{fig:BBH plot}, we replaced the MMLU-Pro benchmark with the BBH benchmark and in \Cref{fig:8B plot}, we used a smaller 8B-parameter model of the same model family instead of the 70B-parameter model.

\subsection{BBH Benchmark}
\subsection{MMLU-Pro Benchmark with 8B-Parameter Model}

\newpage
\section{Acknowledgment of AI Usage}
\label{sec:ai-usage}

\aiProjectName{\scriptsize Multi-Agent Reasoning Improves Compute Efficiency: Pareto-Optimal Test-Time Scaling}
\aiDomain{Paper}
\aiKeyApplication{Test-Time-Compute Efficiency}

\aiContactName{Florian Valentin Wunderlich}
\aiContactEmail{florian.wunderlich@uni-goettingen.de}
\aiContactAffiliation{University Göttingen} 

\aiModels{ChatGPT (4o, 5), Claude (3.7, 4.0), Gemini (2.5)}

\aiFindingLiterature{Asta Paperfinder}
\aiFindingExamples{Asta Paperfinder}

\aiImprovingContent{ChatGPT}

\aiGeneratingCode{ChatGPT, Claude, Gemini}
\aiRefactoringCode{ChatGPT, Claude, Gemini}

\aiWhyUse{Efficiency / Speed\\Expertise Access}
\aiMitigateErrors{We manually verify all AI-generated content}
\aiMinimizeHarm{We deactivated automatic code and shell execution tools.}

\makeAIUsageCard

\clearpage
\onecolumn
\hypertarget{annotation}{}
\pagestyle{empty}
\lstset{
  basicstyle=\footnotesize\ttfamily,
  breaklines=true,
  breakatwhitespace=false,
  columns=flexible,
  numbers=none
}

\definecolor{Primary}{RGB}{59, 130, 246}    %
\definecolor{PrimaryDark}{RGB}{30, 64, 175} %
\definecolor{LightBg}{RGB}{239, 246, 255}   %
\definecolor{TextDark}{RGB}{31, 41, 55}     %
\definecolor{TextMuted}{RGB}{107, 114, 128} %

\begin{tikzpicture}[remember picture, overlay]
  \fill[Primary] ([xshift=0cm,yshift=0cm]current page.north west) rectangle ([xshift=\paperwidth,yshift=-0.4cm]current page.north west);
\end{tikzpicture}

\vspace{0.8cm}
\begin{center}
  {\fontsize{22}{26}\selectfont\sffamily\bfseries \textcolor{PrimaryDark}{CiteAssist}}\\[0.2em]
  {\Large\sffamily\scshape \textcolor{TextMuted}{Citation Sheet}}\\[0.8em]
  {\small\sffamily Generated with \href{https://citeassist.uni-goettingen.de/}{\textcolor{Primary}{\texttt{citeassist.uni-goettingen.de}}}
  \CiteAssistCite{}
  }\end{center}

\begin{center}
\vspace{1em}
\begin{tikzpicture}
\draw[Primary, line width=0.6pt] (0,0) -- (\textwidth,0);
\end{tikzpicture}
\vspace{1.2em}
\end{center}

\begin{tcolorbox}[enhanced,
                 frame hidden,
                 boxrule=0pt,
                 borderline west={2pt}{0pt}{Primary},
                 colback=LightBg,
                 sharp corners,
                 breakable,
                 fonttitle=\sffamily\bfseries\large,
                 coltitle=Primary,
                 title=BibTeX Entry,
                 attach title to upper={\vspace{0.2em}\par},
                 left=12pt]
\lstset{
    inputencoding = utf8,  %
    extendedchars = true,  %
    literate      =        %
      {á}{{\'a}}1  {é}{{\'e}}1  {í}{{\'i}}1 {ó}{{\'o}}1  {ú}{{\'u}}1
      {Á}{{\'A}}1  {É}{{\'E}}1  {Í}{{\'I}}1 {Ó}{{\'O}}1  {Ú}{{\'U}}1
      {à}{{\`a}}1  {è}{{\`e}}1  {ì}{{\`i}}1 {ò}{{\`o}}1  {ù}{{\`u}}1
      {À}{{\`A}}1  {È}{{\`E}}1  {Ì}{{\`I}}1 {Ò}{{\`O}}1  {Ù}{{\`U}}1
      {ä}{{\"a}}1  {ë}{{\"e}}1  {ï}{{\"i}}1 {ö}{{\"o}}1  {ü}{{\"u}}1
      {Ä}{{\"A}}1  {Ë}{{\"E}}1  {Ï}{{\"I}}1 {Ö}{{\"O}}1  {Ü}{{\"U}}1
      {â}{{\^a}}1  {ê}{{\^e}}1  {î}{{\^i}}1 {ô}{{\^o}}1  {û}{{\^u}}1
      {Â}{{\^A}}1  {Ê}{{\^E}}1  {Î}{{\^I}}1 {Ô}{{\^O}}1  {Û}{{\^U}}1
      {œ}{{\oe}}1  {Œ}{{\OE}}1  {æ}{{\ae}}1 {Æ}{{\AE}}1  {ß}{{\ss}}1
      {ẞ}{{\SS}}1  {ç}{{\c{c}}}1 {Ç}{{\c{C}}}1 {ø}{{\o}}1  {Ø}{{\O}}1
      {å}{{\aa}}1  {Å}{{\AA}}1  {ã}{{\~a}}1  {õ}{{\~o}}1 {Ã}{{\~A}}1
      {Õ}{{\~O}}1  {ñ}{{\~n}}1  {Ñ}{{\~N}}1  {¿}{{?\`}}1  {¡}{{!\`}}1
      {„}{\quotedblbase}1 {“}{\textquotedblleft}1 {–}{$-$}1
      {°}{{\textdegree}}1 {º}{{\textordmasculine}}1 {ª}{{\textordfeminine}}1
      {£}{{\pounds}}1  {©}{{\copyright}}1  {®}{{\textregistered}}1
      {«}{{\guillemotleft}}1  {»}{{\guillemotright}}1  {Ð}{{\DH}}1  {ð}{{\dh}}1
      {Ý}{{\'Y}}1    {ý}{{\'y}}1    {Þ}{{\TH}}1    {þ}{{\th}}1    {Ă}{{\u{A}}}1
      {ă}{{\u{a}}}1  {Ą}{{\k{A}}}1  {ą}{{\k{a}}}1  {Ć}{{\'C}}1    {ć}{{\'c}}1
      {Č}{{\v{C}}}1  {č}{{\v{c}}}1  {Ď}{{\v{D}}}1  {ď}{{\v{d}}}1  {Đ}{{\DJ}}1
      {đ}{{\dj}}1    {Ė}{{\.{E}}}1  {ė}{{\.{e}}}1  {Ę}{{\k{E}}}1  {ę}{{\k{e}}}1
      {Ě}{{\v{E}}}1  {ě}{{\v{e}}}1  {Ğ}{{\u{G}}}1  {ğ}{{\u{g}}}1  {Ĩ}{{\~I}}1
      {ĩ}{{\~\i}}1   {Į}{{\k{I}}}1  {į}{{\k{i}}}1  {İ}{{\.{I}}}1  {ı}{{\i}}1
      {Ĺ}{{\'L}}1    {ĺ}{{\'l}}1    {Ľ}{{\v{L}}}1  {ľ}{{\v{l}}}1  {Ł}{{\L{}}}1
      {ł}{{\l{}}}1   {Ń}{{\'N}}1    {ń}{{\'n}}1    {Ň}{{\v{N}}}1  {ň}{{\v{n}}}1
      {Ő}{{\H{O}}}1  {ő}{{\H{o}}}1  {Ŕ}{{\'{R}}}1  {ŕ}{{\'{r}}}1  {Ř}{{\v{R}}}1
      {ř}{{\v{r}}}1  {Ś}{{\'S}}1    {ś}{{\'s}}1    {Ş}{{\c{S}}}1  {ş}{{\c{s}}}1
      {Š}{{\v{S}}}1  {š}{{\v{s}}}1  {Ť}{{\v{T}}}1  {ť}{{\v{t}}}1  {Ũ}{{\~U}}1
      {ũ}{{\~u}}1    {Ū}{{\={U}}}1  {ū}{{\={u}}}1  {Ů}{{\r{U}}}1  {ů}{{\r{u}}}1
      {Ű}{{\H{U}}}1  {ű}{{\H{u}}}1  {Ų}{{\k{U}}}1  {ų}{{\k{u}}}1  {Ź}{{\'Z}}1
      {ź}{{\'z}}1    {Ż}{{\.Z}}1    {ż}{{\.z}}1    {Ž}{{\v{Z}}}1  {ž}{{\v{z}}}1
  }
\begin{lstlisting}
@inproceedings{wunderlich2026,
  author={Wunderlich, Florian Valentin and Kaesberg, Lars Benedikt and Wahle, Jan Philip and Ruas, Terry and Gipp, Bela},
  title={Multi-Agent Reasoning Improves Compute Efficiency: Pareto-Optimal Test-Time Scaling},
  pages={14},
  year={2026},
  month={04}
}
\end{lstlisting}
\end{tcolorbox}

\vfill
\begin{tikzpicture}
\draw[Primary!40, line width=0.4pt] (0,0) -- (\textwidth,0);
\end{tikzpicture}
\begin{center}
\small\sffamily\textcolor{TextMuted}{Generated \today}
\end{center}

\twocolumn

\end{document}